\documentclass[runningheads]{llncs}
\usepackage{amsmath}

\usepackage{graphicx}
\usepackage[table,xcdraw]{xcolor}
\usepackage{cite}
\usepackage{amssymb,amsfonts}
\usepackage{import}
\usepackage{xcolor}
\usepackage{algorithm}  
\usepackage{algorithmic}

\usepackage{booktabs}

\usepackage{times}
\usepackage{soul}
\usepackage{url}
\usepackage[hidelinks]{hyperref}
\usepackage[utf8]{inputenc}
\usepackage[small]{caption}
\usepackage{textcomp}
\usepackage[table,xcdraw]{xcolor}
\usepackage{multirow}

\begin{document}

\title{KAE: A Property-based Method for Knowledge Graph Alignment and Extension
}


\author{Daqian Shi, Xiaoyue Li and Fausto Giunchiglia}
\institute{Department of Information Engineering and Computer Science (DISI),\\ University of Trento, Italy\\
\email{\{daqian.shi, xiaoyue.li, fausto.giunchiglia\}@unitn.it}
}

\authorrunning{D. Shi, X. Li and F. Giunchiglia}

\maketitle

\begin{abstract}
A common solution to the semantic heterogeneity problem is to perform knowledge graph (KG) extension exploiting the information encoded in one or more candidate KGs, where the alignment between the reference KG and candidate KGs is considered the critical procedure. However, existing KG alignment methods mainly rely on entity type (etype) label matching as a prerequisite, which is poorly performing in practice or not applicable in some cases. In this paper, we design a machine learning-based framework for KG extension, including an alternative novel property-based alignment approach that allows aligning etypes on the basis of the properties used to define them. The main intuition is that it is properties that intentionally define the etype, and this definition is independent of the specific label used to name an etype, and of the specific hierarchical schema of KGs. 
Compared with the state-of-the-art, the experimental results show the validity of the KG alignment approach and the superiority of the proposed KG extension framework, both quantitatively and qualitatively.

\end{abstract}

 \section{Introduction}
The semantic heterogeneity problem arises whenever there is a need to exploit knowledge graphs (KGs) from heterogeneous resources \cite{lonsdale2010reusing}. Here we focus on KGs where nodes are entities decorated by data properties and relations are object properties. Furthermore, we talk of entity type (etype) meaning the class to which an entity belongs, e.g., \textit{Person} and \textit{Event}. A solution of semantic heterogeneity is to perform \textit{KG extension} that extends the reference KG by knowledge encoded in one or more candidate KGs. Meanwhile, KG alignment is an attractive topic that involves various tasks, e.g., well-known ontology matching (OM), where two practical tasks are considered critical steps to achieving KG extension. Firstly, the alignments between etypes from reference and candidate KGs  \cite{euzenat2007ontology,giunchiglia2004s,algergawy2018results}, exploiting only schema level information, as it is the most often case in OM. Secondly, the alignments between reference etypes and candidate entities, exploiting additional information of entities, also known as etype recognition \cite{portisch2021background,shalaby2016entity}. These two kinds of alignments are prerequisites for extending the reference KG at both the schema level and instance level, respectively. 

Etype alignment is a prepositive task during KG extension since etypes define the schema of collected entities in a KG. The reference KG can directly integrate candidate entities if their corresponding etypes align with reference etypes. Existing etype alignment methods mainly exploit natural language processing (NLP) based \cite{cheatham2013string,sun2015comparative} and structure-based \cite{jimenez2011logmap,faria2013agreementmakerlight} techniques. Both techniques enforce etype label matching as a prerequisite. NLP-based methods utilize diverse lexical-based similarity metrics and synonym analysis to align etype labels, which raises limitations when applied in practice. Labels may suggest a wrong etype \cite{sleeman2015entity}, where the same concept can be labeled differently by KGs. For instance, an eagle can be labeled as \textit{Bird} in a general-purpose KG and \textit{Eagle} in a domain-specific KG. In turn, the same label may present different concepts in heterogeneous KGs, which will also lead to wrong recognition results. Structure-based methods consider the KG hierarchy as an additional input, where the structure of a hierarchy is used to drive the label matching, e.g., it is suggested to perform label matching to sub-classes or super-classes. However, these methods may also mislead the conclusions as properties assigned to an etype in the hierarchy are cumulative and depend only on nodes in the path from the root and, therefore, do not depend on the order by which they are assigned \cite{giunchiglia2020entity}. In addition, the difference in taxonomy between KGs will increase the impact of such mistakes, e.g., the super-class of etype \textit{Eagle} can be \textit{Animal} in one KG and \textit{Bird} in another KG. On the other hand, the alignments between etypes and entities (here we call etype recognition task) are also critical since the reference KG can also be extended by candidate entities that are not included in aligned etypes. However, it is not easy to identify such alignments due to the lexical labels are not applicable to alignment entities and etypes. For instance, entity \textit{apple} can be a company, a fruit, or the name of a pet, but there is only trivial lexical similarity between entity \textit{apple} and etypes \textit{company}, \textit{fruit} and \textit{pet}. Thus, current NLP-based and structure-based methods poorly perform the etype recognition task either. 

As a solution to the above problems, the main intuition of this paper is to align etypes/entities with reference etypes for KG extension on the basis of the properties used to define them. It is, in fact, the properties that are used to intentionally define an etype, and this definition is independent of the specific label and also independently of its hierarchy \cite{ganter2012formal}. This allows us to think of etypes as being organized in hierarchies, where lower etypes inherit properties from upper etypes and where the entities populating an etype also populate all the upper etypes. In turn, it allows us to think of KG alignment as a problem of matching inheritance hierarchies, where etypes may or may not be populated with entities. Figure \ref{fig:1} (a) provides one such hierarchy, used as a running example in the rest of the paper. In practice, most relevant KGs are associated with large numbers of properties, see, e.g., DBpedia \cite{auer2007dbpedia} and OpenCyc \cite{farber2015comparative}. And the reason for this is quite obvious, being that one of the purposes of KG extension is exactly that of extending the number of properties. 

In this paper, we propose a machine learning (ML)-based framework to realize KG extension task, including an algorithm for generic KG alignment based on the above intuition of exploiting properties. We introduce a formalization strategy, where we organize a KG and its inner mappings between etypes/entities and properties based on the use of formal concept analysis (FCA) \cite{ganter2012formal}. Then, we present three \textit{property-based} metrics to measure the similarity between etypes and entities, where the metrics characterize the role that properties have in the definition of given etypes from different aspects. They capture the main idea that the number of aligned properties affects the contextual similarity between etypes and entities. The proposed similarity metrics and algorithm are applied in two key modules of the KG extension framework, i.e., etype alignment and KG extension modules. Finally, we will obtain an extended reference KG by integrating etypes and entities from the candidate KGs.

Overall, the main contributions of this paper are as follows:
\begin{itemize}
    \item We design an ML-based framework for KG alignment and extension tasks, where etype alignment and etype recognition are introduced as two key procedures. 
    \item We proposed a novel set of property-based metrics for measuring contextual similarity between KGs while introducing an FCA-based KG formalization strategy.
    \item We implement a classification-based method for KG alignment, which exploits the property-based information by our proposed metrics.
    \item We compare our method with state-of-the-art etype recognition methods from several different perspectives. The experimental results show the validity of the similarity metrics and the superiority of the proposed KG extension framework, both quantitatively and qualitatively.
\end{itemize}

The rest of the paper is organized as follows. Section 2 discusses the motivation for exploiting properties for etype recognition and section 3 demonstrates how we formalize a KG and its inner mappings into FCA contexts. Section 4 introduces three specificity measurements and their corresponding property-based etype similarity metrics. In section 5, we describe the proposed ML-based KG extension framework and the details of individual modules. Then we discuss the experiment setups and experimental results in Section 6. Finally, we present the related work in Section 7 and conclude the paper in Section 8.

\section{Intuitive Discussion}
\label{sec2}
\subsection{Task Description}
In the knowledge integration area, KG alignment and extension can be organized as sequential tasks, where KG alignment aims to align the knowledge from candidate KG and reference KG, and KG extension aims to integrate target information into the reference KG following the aligned knowledge. To clearly define the above-mentioned tasks, we define a KG as a hierarchy of concepts, where properties are used to describe them. Specifically, we define the schema of a KG and its inner relations as $KGS = \langle C,P,R \rangle$, where $C = \{C_1,...,C_n\}$ being the classes of entities (i.e. etypes), $P = \{p_1,...,p_m\}$ being the set of properties, $R = \{\langle C_i,T(C_i) \rangle|C_i \in C \}$ being the set of correspondences between etypes and properties, and function $T(C_i)$ returns properties associated with $C_i$. As for the entities $I$, we define $I = \{I_1,...,I_l\}$ being the set of entities in $KG$, where each entity $I_i$ can be identified by one or several etypes. $t(I_i)$ refers to the set of associations between entities and properties. We consider that the property $p_i$ is used to describe an etype $C_i$ or an entity $I_i$ when the property belongs to set $T(C_i)$ or $t(I_i)$, respectively. Thus, given a reference $KG_{ref}$ and a candidate $KG_{cand}$, considering the two cases in KG alignment, etype alignment task aims to align the candidate etypes $C_{cand}$ with the reference etypes $C_{ref}$, and etype recognition task aims to align the candidate entities $I_{cand}$ with $C_{ref}$, where $C_{cand}, I_{cand} \in KG_{cand}$, $C_{ref} \in KG_{ref}$\footnote{Notice both etype alignment and etype recognition are specific tasks of KG alignment.}.

\subsection{Observations and Motivations}

\begin{figure}[!t]
	\centering
	\setlength{\abovecaptionskip}{5pt}    
    \setlength{\belowcaptionskip}{1pt}
	\includegraphics[width=0.99\linewidth]{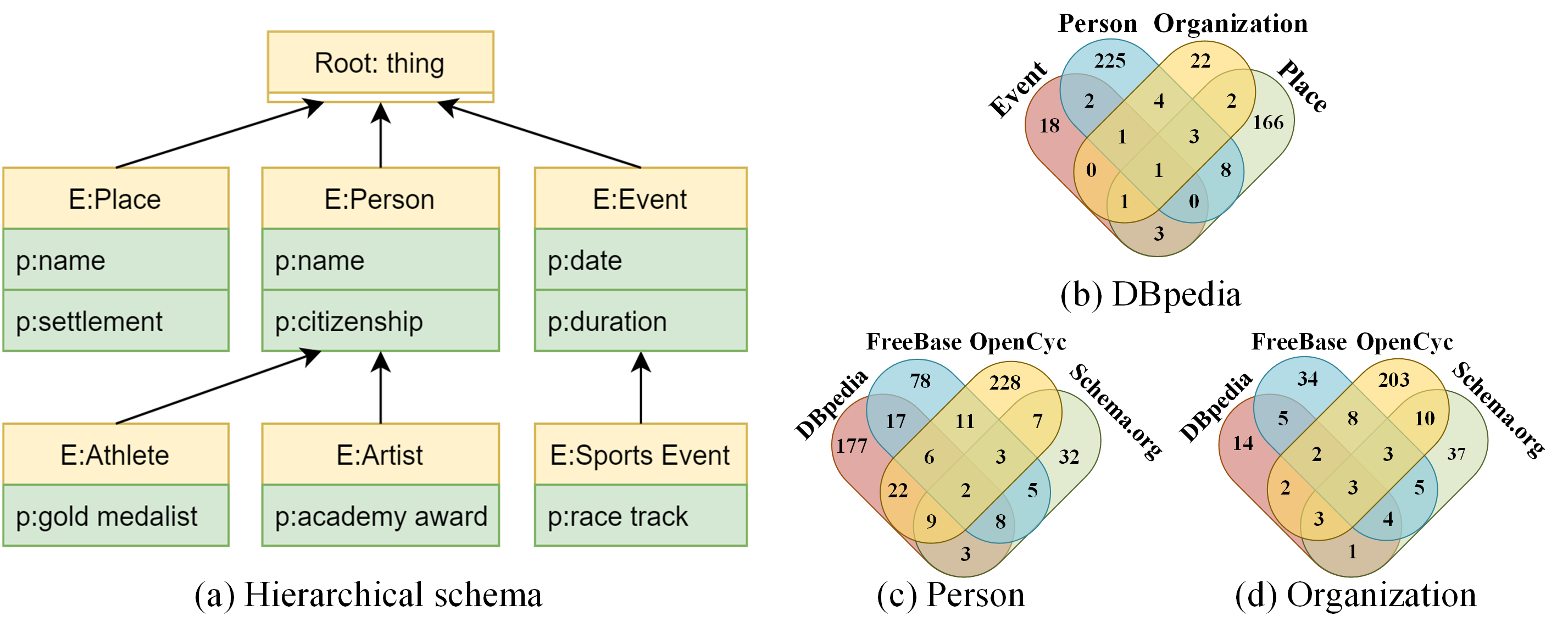}
	\caption{An example of the hierarchical schema in KG.
	\label{fig:1}}
\end{figure}

Property is one of the most basic and critical elements for intentionally defining KG concepts, which is independent of the specific label used to name concepts and of the specific hierarchical schema of KGs \cite{giunchiglia2021property}. For each KG schema, etypes play the role of categorization, and properties aim to draw sharp lines so that each entity in the domain falls determinately either in or out of each etype \cite{giunchiglia2019knowledge}. Meanwhile, we also have the following observations when comparing properties across different KGs: 
\begin{itemize}
    \item In a specific KG, each etype is described by a set of properties, whereas most of the properties are distinguishable according to the belonging etypes and a small number of properties are shared across different etypes.
    \item  Same or similar properties are shared across different KGs for describing same concepts.
\end{itemize}
To clearly present these observations, we introduce a special type of \textit{Venn graphs}, namely \textit{knowledge lotuses}, to represent the shareability of properties that occurs within and across KGs. Knowledge lotuses provide a synthetic view of how different KGs overlap in properties \cite{giunchiglia2020entity}. Figure \ref{fig:1} (b)-(d) show several examples\footnote{The values in the lotuses in Figure \ref{fig:1} (b)-(d) are computed via a lexical analysis method.}, where we assume that we have four contexts built from (parts of) the four biggest KGs, namely OpenCyc\footnote{www.cyc.com}, DBpedia\footnote{https://www.dbpedia.org/}, Schema.org\footnote{https://schema.org/}, and FreeBase \cite{berant2013semantic}. Each value in a lotus refers to the number of shared properties. For instance, in Figure \ref{fig:1} (b), we notice that the two etypes \textit{Person} and \textit{Organization} share four properties in DBpedia. In Figure \ref{fig:1} (c), we can see that the definitions of \textit{Person}, as in OpenCyc and DBpedia, share twenty-three properties. Meanwhile, we also present concrete shared properties of etype \textit{Person} across different KGs in Table \ref{PropertyExample}, e.g., \textit{birth} and \textit{education} are applied in both OpenCyc and DBpedia. All these examples present features of properties, namely, the unity for describing the same concept and the diversity for distinguishing different concepts. For instance, a \textit{Person} can be distinguished from a \textit{Place} by the property \textit{birth}, which is a crucial step to align etypes and identify the type of entities. Thus, inspired by these observations, we find that it is important to exploit and measure properties for better performance of KG alignment and extension tasks.

\begin{table*}[!t]
\centering
\caption{Shared Properties of etype \textit{Person} across different KGs.}
  \label{PropertyExample}

\begin{tabular}{@{}lcccl@{}}
\toprule
\textbf{Contexts}      & & \textbf{Tot.} &  & \textbf{Shared Properties} \\  \midrule
OpenCyc \& DBpedia     & & 39  & & \textit{birth, education, title, activity, ethnicity, employer, status...} \\
OpenCyc \& Schema.org  & & 21  & & \textit{contact, suffix, tax, job, children, works, worth, gender, net...} \\
DBpedia \& FreeBase    & & 33  & & \textit{title, number, related, birth, parent, work, name...} \\ 
DBpedia \& Schema.org  & & 22  & & \textit{death, sibling, point, member, nationality, award, parents...} \\ 
\bottomrule                    
\end{tabular}
\end{table*}

\section{Knowledge Graph Formalization}
\label{sec3}
We formalize the relation between properties and KG concepts as \textit{associations} to introduce the property information into the target KG tasks. At the schema level, the KG schema will be flattened into a set of triples, where each triple encodes information about \textit{etype-property} associations, e.g., triple “organization-domain-LocatedIn” encodes the “organization-LocatedIn” association. Instance-level cases generally define triples as “entity-property-entity", where two associations are encoded. For instance, instance-level triple “EiffelTower-LocatedIn-Paris” encodes two \textit{entity-property} associations “EiffelTower-LocatedIn" and “LocatedIn-Paris". We introduce FCA \cite{ganter2012formal} to encode such associations as a formalization of the KGs we process. Notice that both the schema-level and instance-level associations are included in the formalization. Specifically, we have two working assumptions:
\begin{itemize}
    \item We consider both etypes $C$ and entities $I$. Similar to general formalization methods \cite{stumme2001fca,cure2015formal}, we associate an entity with its set of properties $t(I_i)$. Different from general methods, we also associate an etype with its set of properties $T(C_i)$. 
    \item Etype characterization exploits not only the properties associated with it but also the properties which are associated with other concepts. Thus, we introduce the notion of \textit{unassociated} properties and exploit this distinction in the formalization process. 
\end{itemize}

\begin{figure*}[!t]
	\centering
	\includegraphics[width=1\linewidth]{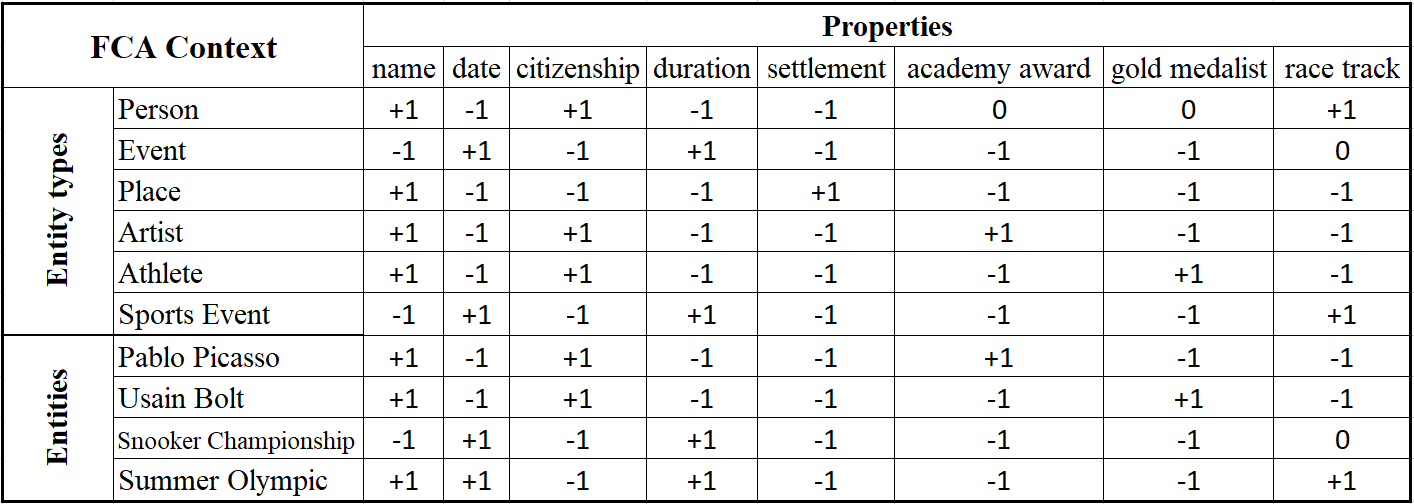}
	\caption{An example of formalizing KG into FCA contexts 
	\label{fig:2}}
\end{figure*}

\noindent
As an example, the hierarchy of the KG schema in Figure \ref{fig:1} is extracted from DBpedia \cite{auer2007dbpedia}. In each box, etypes are presented in yellow and their properties in green. The arrow refers to the \textit{sub-class of} relation between two etypes. We then formalize these etypes into an FCA context as shown in Figure \ref{fig:2}. We adopt the following conventions. The value box with a “+1” represents the fact the property is associated with the etype, e.g., \textit{citizenship} is associated with \textit{Person}. The value box with a “-1” means the property is unassociated with the etype, e.g., \textit{date} is not used to describe etype \textit{Person}. The value “0” (for undefined) represents the fact that the property is unassociated with the etype but associated with one of its sub-classes. The intuition is that the property might or might not be used to describe the current etype, e.g., \textit{academy award} is used to describe \textit{Artist} and it might be used to describe \textit{Person} since \textit{Artist} is a subclass of \textit{Person}. Similar to etypes, we can also find formalized entities and their properties in Figure \ref{fig:2}. Need to notice that these entities are selected from the KG with a hierarchical schema, thus, they can inherit the \textit{unassociated properties} from their etypes, e.g., as an \textit{Athlete}, \textit{Usain Bolt} does not have property \textit{duration}. 

We encode the above-mentioned three correlations as the parameter $w_E(p)$. Since the correlation of “associated with” is positive for a property-based description, the correlation of “unassociated with” is negative and the correlation of “undefined” is neutral, we define the parameter as:
\begin{equation}
w_E(p) = \left\{
\begin{tabular}{ll}
1,  & if $p \in prop(E)$\\
0,  & if $p \notin prop(E) \land p \in prop(E.subclass)$ \\
-1, & if $p \notin prop(E) \land p \notin prop(E.subclass)$ 
\end{tabular}\right.
\label{equ:1}
\end{equation}
where we suppose $K$ as the union of etypes $C$ and entities $I$ in a KG for a general demonstration and the etype/entity $E \in K$, $p$ is the target property, $E.subclass$ refers to the sub-classes of the etype $E$ and $prop(E)$ refers to the properties associated with $E$. Need to notice a special case that \textit{undefined properties} also exist where a specific entity misses the inherited property, such as \textit{race track} is used to describe \textit{Sports events} but missed for its entity \textit{Snooker Championship}, which will also make $w_E(p)=0$. In this case, we encode properties for each etype and then record the properties associated with each entity. If an entity does not contain a property it should inherit from its etype, it is considered undefined. 
With our designed KG formalization method, we distinguish several concrete situations for learning the association between properties and etypes/entities and encoding more property information. The proposed formalized approach serves as a pre-process for subsequent property-based similarity methods used in computations in the following sections.


\section{Property-based Similarity Metrics}
\label{sec4}
One of the intuitions of our work is to identify etypes and entities by properties that are essential elements for defining KG concepts \cite{giunchiglia2020entity}. The reference etype tends to match with the candidate etype/entity when they have properties overlapped. Therefore, it is critical to measure the overlapped properties between etypes since etypes with completely overlapped properties rarely happen in real-world KGs \cite{fumagalli2021ranking}. Meanwhile, we also exploit the intuitions underlying the normalization of the “get-specific” heuristic provided in \cite{giunchiglia2007formalizing} to distinguish the weights of different overlapped properties. The key inspiration is that properties at different levels of specificity have different relevance in the etype recognition. In particular, a more specific property provides more information that allows for defining concepts. As a result, in this section, we introduce three notions of \textit{horizontal specificity}, \textit{vertical specificity}, and \textit{informational specificity} and their corresponding similarity metrics for measuring the degree of overlapped properties.

\subsection{Horizontal Specificity}

For measuring the specificity of a property, a possible idea is to horizontally compare the number of etypes that are described by a specific property, namely the shareability of the property \cite{giunchiglia2020entity}. If a property is used to describe diverse etypes, it means that the property is not highly characterizing its associated etypes. Thus, for instance, in figure \ref{fig:1}, the property \textit{name} is used to describe \textit{Person}, \textit{Place}, \textit{Athlete} and \textit{Artist}, where \textit{name} is a common property that appears in different contexts. Dually,  \textit{settlement} is a horizontally highly specific property since it is  associated only with the etype \textit{Place}. 
Based on this intuition, we consider the specificity of a property as related to its shareability. Therefore, we propose $HS$ (\textit{Horizontal Specificity}) for measuring property specificity. More precisely, $HS$  aims to measure the number of etypes that are associated with the target property in a specific KG. We model $HS$ as: 

\begin{equation}
HS_{KG}(E,p)= w_E(p) * {e^{\lambda(1-|K_p|)}}
\label{equ:2}
\end{equation}
where: $p$ is the input property and $K_p$ is the set of etypes described by the input property in a specific $KG$ and $|K_p|$ is the number of etypes in $K_p$, thus $|K_p| \ge 1$; $e$ denotes the natural mathematical constant \cite{finch2003mathematical}; $\lambda$ represents a constraint factor, the range of $\lambda$ is $(0, 1]$. The reason for using an exponential function to model the $HS$ is that we aim to normalize the horizontal specificity. The motivation is that different properties may have a larger difference on $|K_p|$ in large KGs. Need to note that the range of the $HS$ is $[-1, 1]$, and the extremum points both indicate that the property is shared by one or a few etypes, signifying that the property is highly specific. The impact of such specificity is determined by the target etypes used for description, which is captured by the $\lambda$, resulting in different outcomes. 

\subsection{Vertical Specificity}

Etypes are organized into classification hierarchies such that the upper-layer etypes represent more abstract or more general concepts, whereas the lower-layer etypes represent more concrete or more specific concepts \cite{giunchiglia2007formalizing,rios2013learning}. Correspondingly, properties of upper-layer etypes are more general since they are used to describe general concepts, vice versa, properties of lower-layer etypes are more specific since they are used to describe specific concepts. We assume that lower-layer properties will contribute more to the identification of an etype since they are more specific. For instance, in Figure \ref{fig:1}, as a lower-layer etype, \textit{Artist} can be identified by the property \textit{academy award} but not by the property \textit{name}. Based on this intuition, we propose $VS$ for capturing the vertical specificity, as follows:

\begin{equation}
VS_{KG}(E,p)= w_E(p) * \min_{E \in K_p} layer(E)
\label{equ:3}
\end{equation}
where $layer(E)$ refers to the calculation based on the layer of the inheritance hierarchy where an etype $E$ is defined, under a min-max normalization. Note that $layer(E) \in [0,1]$ and all $E$ in set $K_p$ are described by $p$. The range of $VS$ is $[-1,1]$, and extremum of the $VS = 1$ and $VS = -1$ demonstrates the property is hierarchically specific while holding opposing impacts on the target etype, and $VS =  0$ demonstrates the property is not specific enough from the hierarchical point of view. 

\subsection{Informational Specificity}
Horizontal specificity allows measuring the shareability of properties, which is independent and does not change (increase/decrease) with the number of entities populating it.  We take into account this fact by introducing the notion of informational specificity $IS$. The intuition is that $IS$ will decrease when the entity counting increases. Thus, for instance, the $IS$ of \textit{gold medalist} decreases when there are increasing entities of \textit{athletes}, as from the schema in Figure \ref{fig:2}. Clearly, $IS$, differently from $HS$, can be used in the presence of entities. The definition of informational specificity is inspired by Kullback–Leibler divergence theory \cite{van2014renyi}, which is introduced to measure the difference between two sample distributions $Y$ and $\hat{Y}$. More specifically, given a known sample distribution $Y$, assume that a new coming attribute $x$ changes $Y$ to $\hat{Y}$. Then, the Kullback–Leibler divergence theory demonstrates that the importance of $x$ for defining $Y$ is positively related to the difference between $Y$ and $\hat{Y}$. In the definition of informational specificity, we need to exploit some notions from information theory, where we apply informational entropy $H(K)$ as: 

\begin{equation}
H(K_v)= \frac{- \sum_{i=1}^{|K_v|} F(E_i)\log\frac{F(E_i)}{F(K_v)}}{F(K_v)}
\label{equ:4}
\end{equation}
where $K_v$ refers to any subset of $K$ in a KG and $|K_v|$ is the number of etypes included; $H(\cdot)$ represents the informational entropy of an etype set; $E_i$ is a specific etype in set $K_v$, thus $|K_v| \ge 1$; $F(E_i)$ refers to the number of samples of etype $E_i$, and $F(K_v)$ refers to the number of all samples in $K_v$. 
Need to notice when we calculate the informational entropy for KGs without entities, $F(E_i) = 1$ since each KG includes one etype sample. For KGs with entities, $F(E_i)$ and $F(K_v)$ depend on the number of samples of the given etype and the subset of KG. After obtaining informational entropy, the informational specificity $IS$ of property is defined as:
\begin{equation}
IS_{KG}(E,p)= w_E(p) * (H(K) -  H(\hat{K}))
\label{equ:5}
\end{equation}
where $K$ is the set of all etypes in $KG$, and $\hat{K}$ is the set of etypes without associated with the input property $p$, thus $\hat{K} = K - K_p$. Being subtracted by the overall informational entropy $H(K)$, $IS$ presents the importance of the property $p$ for describing the given etype set $K$. For computational convenience, we employ the min-max normalization method to constrain the calculation results within the range of $[-1, 1]$.  

\subsection{Similarity Metrics}
We have modeled the specificity of properties, which represent their weights for describing KGs from different measuring aspects. Then, we define three similarity metrics based on the corresponding specificity to measure the property overlapping between two concepts. Given two KGs, the reference KG $A$ and candidate KG $B$, we define a function for calculating different similarities between etypes/entities from $A$ and $B$ based on their corresponding specificity: 
\begin{equation}
Sim(E_a,E_b) =\frac{1}{2}\sum_{i=1}^{k} \left ( \frac{SPC_{A}(E_a,p_i)}{|prop(E_a)|} + \frac{SPC_{B}(E_b,p_i)}{|prop(E_b)|} \right)
\label{equ:6}
\end{equation}

\noindent where we take $E_a$, $E_b$ as the candidate etypes from $A$ and $B$ respectively, thus $E_a \in A$ and $E_b \in B$; $prop(E)$ refers to the properties associated with the specific etype and $|prop(E)|$ is the number of properties in $prop(E)$; $SPC_{ETG}(\cdot)$ represents the specificity measurements we defined above, $SPC(\cdot) = \{HS(\cdot),VS(\cdot),IS(\cdot) \}$, thus $SPC_{A}(E_a,p_i)$ and $SPC_{B}(E_b,p_i)$ refer to the specificity of the aligned property $p_i$ in $A$ and $B$ respectively; $k$ is the number of aligned properties which are associated with both etype $E_a$ and $E_b$. As a result, we obtain three similarity metrics which are horizontal similarity $Sim_H$, vertical similarity $Sim_V$ and informational similarity $Sim_I$. 

Notice that each similarity metric is symmetric, more specifically, $Sim(E_a,E_b) = Sim(E_b,E_a)$. Note also that we apply z-score normalization \cite{patro2015normalization} to all similarity metrics at during calculations to constrain the range of $Sim_H, Sim_V ,Sim_I$ in $[0, 1]$ for computational convenience purposes. The normalization following the function $z = \frac{x - \mu}{\sigma}$, where $z$ is the normalized value, $x$ is the original value, $\mu$ is the mean of all values, and $\sigma$ is the standard deviation of all values.

\section{The Proposed Method}
\label{sec5}
\begin{figure*}[!t]
	\centering
	\includegraphics[width=0.6\linewidth]{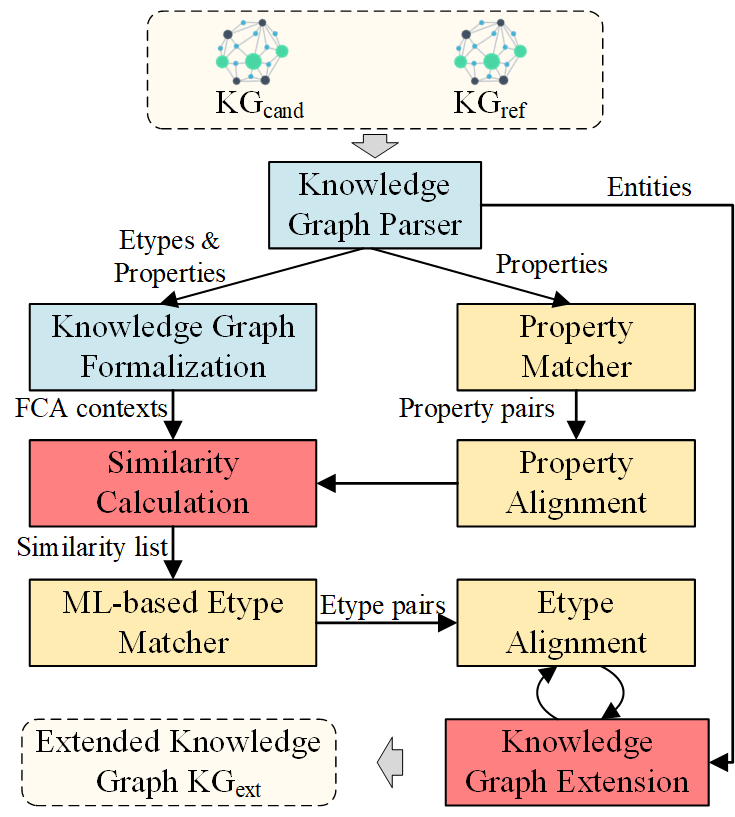}
	\caption{The framework of KG alignment and extension. 
	\label{fig:3}}
\end{figure*}

In order to extend the reference KG by given candidate KGs, we propose a framework that exploits the FCA formalization and property-based similarity metrics defined above, shown in Figure \ref{fig:3}. The framework mainly consists of six modules, namely: KG parser, KG formalization module, property matcher, similarity calculation module, etype matcher, and KG extension module. Notice that modules are marked in different colors for distinguishing their usage. 

Different from KG completion task \cite{zhao2020connecting,yaghoobzadeh2017multi} which manipulates entities in a single KG, there will be two (or more) KGs involves in the KG extension task, where we have $KG_{ref}$ and $KG_{cand}$ as inputs. The KG parser aims to parse the input KG as a structured set of etypes, entities, and properties. The KG will be flattened into a set of triples, where associations will be extracted from triples. Then the triples and associations will be used to generate an FCA for each input KG by the KG formalization module\footnote{The KG parser and the formalization module are marked in blue since they are data processing modules in the framework.}. After all properties of input KGs have been collected, they are sent to the NLP-based property matcher. Different labels of properties may express similar meanings since many of them are variations of the same label. Thus, an NLP pipeline is designed to normalize all input properties, where phrase segmentation, lemmatization, and stop-word removal are introduced for better normalization performance. String-based and language-based similarity metrics are exploited for matching the properties by normalized labels \cite{bella2017language,bella2016domain}. Then, we align properties from input KGs following the results of property matcher. In the next phase, we generate the proposed property-based similarity metrics $Sim_H$,  $Sim_V$, and $Sim_I$ by inputting the FCA contexts and aligned properties. According to the Function (\ref{equ:6}), three similarity values will be generated for each etype pair, which will then be passed to the etype matcher together with general string-based and language-based similarity metrics. We will align etypes by the output of the etype matcher\footnote{Two matchers and the following alignment operations are marked in yellow.}. In the final step, the knowledge extension module\footnote{Notice the modules with algorithms, i.e., similarity calculation and KG extension module, are marked in red.} integrates entities from the candidate KG according to the new reference KG schema that is updated with aligned etypes. In Figure \ref{fig:3} we use  a circle arrow to demonstrate this process since our method will first implement etype-etype alignment to integrate the aligned etypes directly into the reference KG, and then implement entity-etype alignment to integrate the rest of entities. The final output of the framework is an extended reference KG with integrated etypes and entities.

\subsection{Similarity Calculation Algorithm}

\begin{algorithm}[!t] 
\caption{Calculating horizontal similarity $Sim_H$ between reference and candidate KGs. $L_H=SimilarityCalculation(f_a,f_b)$} 
\label{alg:1} 
\begin{algorithmic}[1] 
\REQUIRE ~~\\ 
Reference and candidate FCA contexts $f_a$, $f_b$;\\
\ENSURE ~~\\ 
List of all horizontal similarities $L_H$;\\

\STATE 	$PM = (p_a \times p_b) = PropertyMatcher(f_a,f_b)$; \{$PM$ is formed as a set of aligned property pairs, where $p_a \in f_a$, $p_b \in f_b$.\}
\STATE $EM = (E_a \times E_b) = EtypeSelector(f_a,f_b)$; \{etypes $E_a, E_b$ from $f_a, f_b$ are assembled as candidate calculation pairs $EM$.\}
\FOR {all $(E_j,E_k) \in EM$}
\STATE $sim_H(E_j,E_k) = 0$; \{initialize the value of horizontal similarity $sim_H(E_j,E_k)$.\}
\FOR {all $(p_n,p_m) \in PM$}
\IF {$p_n \in prop(E_j) \wedge p_m \in prop(E_k)$}
\STATE $sim_H(E_j,E_k).add(\frac{HS_A(E_j,p_n)}{|prop(E_j)|} {+} \frac{HS_B(E_k,p_m)}{|prop(E_k)|})$; \{calculate the corresponding specificity $HS$ for the similarity $sim_H(E_j,E_k)$.\}

\ENDIF
\ENDFOR
\STATE $L_H.stack(\frac{1}{2} * sim_H(E_j,E_k))$; 
\{stack the value of horizontal similarity to the list $L_H$.\}

\ENDFOR

\RETURN $L_H$
\end{algorithmic}
\end{algorithm}

The property-based similarity calculation is one of the critical parts of this work. We detail the calculation as $SimilarityCalculation(\cdot)$, as shown in Algorithm \ref{alg:1}. After formalizing reference $KG_{ref}$ and candidate $KG_{cand}$, we assume that the two FCA contexts $f_a$ and $f_b$ are generated correspondingly. Then we obtain property matching pairs $PM$ from the property matcher. To calculate the similarity of etypes, we need to generate candidate etypes pairs $EM$ for further processing. For each candidate pair in $EM$, we check their correlated properties and update the specificity values to $Sim_H$,  $Sim_V$ and $Sim_I$ when their properties are aligned. After traversing all the candidate pairs, we obtain a complete etype similarity list $L$ which will be used for training the ML model and aligning candidate etypes. Notice that we present the algorithm for calculating the horizontal similarity $Sim_H$ in algorithm \ref{alg:1}, the metrics vertical similarity $Sim_V$ and informational similarity $Sim_I$ will be calculated by the same algorithm where the only modification is to change $HS_{KG}(\cdot)$ to $VS_{KG}(\cdot)$ and $IS_{KG}(\cdot)$, respectively.

\subsection{Knowledge Graph Extension Algorithm}

\begin{algorithm}[!t] 
\caption{Extending reference KG by integrating a candidate KG. $KG_{ext}=KGExtension(KG_{ref},KG_{cand},EM_{ali})$} 
\label{alg:2} 
\begin{algorithmic}[1] 
\REQUIRE ~~\\ 
Reference and candidate knowledge graphs $KG_{ref}$, $KG_{cand}$;\\
Aligned etype pairs $EM_{ali}$; \\
\ENSURE ~~\\ 
The extended reference knowledge graph $KG_{ext}$;\\

\FOR {all $(E_a,E_b) \in EM_{ali}$}
\STATE $E_a.addProperty(E_b.property)$; \{Merging the properties of etype $E_b$ into $E_a$, where $E_a \in KG_{ref}, E_b \in KG_{cand}$. \}
\STATE $E_a.addEntities(E_b.Entities)$; \{Merging the entities of $E_b$ into $E_a$. \}
\FOR {all $E_{sub} \in E_b.subClass$}
\STATE $E_a.addSubClass(E_{sub})$; \{Merging the etypes $E_{sub}$ and their entities into $E_a$.\}
\ENDFOR
\STATE $KG_{ext}.update(E_a)$; 
\ENDFOR

\STATE 	$E_{ref} = listEtypes(KG_{ref})$; 
\STATE 	$Ent_{cand} = listEntities(KG_{cand})$; 

\FOR {all $Ent_i \in Ent_{cand}$}
\IF {$Ent_i \notin EM_{ali}$}
\STATE $E_n = EtypeRecognizer(Ent_i,E_{ref})$; \{Recognizing the etype $E_n$ of candidate entities $Ent_i$, where $E_n, E_{ref} \in KG_{ref}$.\}
\STATE $E_n.addEntities(Ent_i)$;
\ENDIF

\STATE $KG_{ext}.update(E_n)$
\ENDFOR

\RETURN $KG_{ext}$
\end{algorithmic}
\end{algorithm}

With the help of etype alignment results, we extend the reference KG by integrating the entities from the candidate KG $KG_{cand}$, details are shown in Algorithm \ref{alg:2}. For each etype $E_b$ that is aligned with etype $E_a$ from $KG_{cand}$, we directly add its properties and entities into $E_a$ by $addProperty(\cdot)$ and $addEntities(\cdot)$, respectively. We also consider all subclasses of the aligned etype $E_b$ since the subclass inherit the properties of the etype and will bring new entities. If the subclass $E_{sub}$ of $E_b$ is not aligned with any other etypes in $KG_{ref}$, $E_{sub}$ and its associated properties and entities will be merged into $E_a$. Thus, we integrate candidate entities with $KG_{ref}$ when their etypes are able to align with $KG_{ref}$. Then, the proposed property-based similarity metrics are applied to align the rest of the candidate entities with etype $E_{ref}$ from reference KG, namely function $Etype Recognizer(\cdot)$. If we successfully match an etype $E_n$ with the entity $Ent_i$, $Ent_i$ will be merged into $E_n$, if not, $Ent_i$ will be discarded. Need to notice that depending on the real-world application scenario and topic, the KG engineer will decide if it is needed to integrate the not aligned etypes with their correpsonding entities into $KG_{ref}$. Finally, all changes will be updated to $KG_{ref}$ and we will obtain an extended KG $KG_{ext}$ as the final result of our method.

\subsection{Machine Learning-based Matchers}
According to the Algorithm \ref{alg:2} and Figure \ref{fig:3}, we can find there are three ML-based matchers in our proposed framework. Here we present more details of training and setting these matchers. 

\subsubsection{Etype Matcher}
We develop an ML-based method that deals with etype matching as a binary classification task. The main idea is to predict if two incoming etypes are aligned with each other. For applying this method, a list of candidate pairs are generated by pairing etypes from $KG_{cand}$ and $KG_{ref}$. We will record candidate etype pairs $EM_{ali}$ when the result of classification is ``aligned". The proposed property-based similarity metrics $Sim_H$,  $Sim_V$ and $Sim_I$ are introduced to train the ML models for matching etypes. For better performance, we also exploit label-based and language-based similarity metrics, along with property-based similarity metrics for training the etype matcher.

\subsubsection{Etype Recognizer}
The strategy for developing an etype recognizer is very similar to the etype matcher, where we predict if candidate entities are aligned with target etypes. Thus, we will also generate candidate pairs that consist of entities from $KG_{cand}$ and etypes from $KG_{ref}$. Etypes from $KG_{ref}$ will be outputted as the final recognition results. Compared to the etype matcher, the main difference is that the etype recognizer mainly uses property-based similarity metrics as features for model training since the lexical labels are not applicable to match entities and etypes. Thus, property-based similarity metrics $Sim_H$,  $Sim_V$, and $Sim_I$ are applied for the etype recognizer.

\subsubsection{Property Matcher}
The property matcher aims to align properties between KGs, where label-based and language-based similarity metrics are used for modeling training. The matching strategy is same as etype matcher. It is critical to obtain a powerful property matcher since both etype matcher and etype recognizer are based on the result of property matcher. In this section, we discuss the following solutions to reduce the effect of misaligned properties. 

\begin{itemize}
    \item Use of the formalization parameter $w_E(p)$. As we introduced in section 2, besides “associated” (positive) and “unassociated” (negative) properties, we also defined “undefined” properties (neutral). Since misaligned properties will not be used for similarity calculation, they are treated as “undefined” properties which will not affect the model training and reduce the additional interference. However, additional interference from misaligned properties appears if “unassociated” and “undefined” properties are not distinguished. 
    
    \item Use of similarity metrics. Similar to lexical-based similarity metrics, our property-based similarity metrics also allow to match etypes by soft aligning, even if there are few properties not aligned. This will increase the robustness of our etype recognition approach. 

    \item Use of ML-based models. By learning from the practical data from different resources, ML models will propose a learnable strategy rather than a fixed threshold for determining alignments, which will maximize the use of existing aligned properties and minimize the effect of misaligned properties. 
\end{itemize}

\section{Evaluation}
\label{sec6}

In this section, we aim to evaluate our proposed method by two crucial steps during KG integration, including (1) etype alignment, and (2) etype recognition of entities. Thus, we organize this section as follows. Section 6.1 introduces the experimental setups, including the datasets we used, feature selection, and evaluation strategy. Sections 6.2 and 6.3 present the analysis and quantitative evaluation results, respectively. In section 6.4, we also demonstrate the ablation study to explain the setting of parameters.

\subsection{Experimental Setup}

\subsubsection{Dataset Selection.}
For evaluating the result of etype alignment, we exploit the Ontology Alignment Evaluation Initiative\footnote{http://oaei.ontologymatching.org/2021/} (OAEI) as the main reference for the selection of the etype recognition problems. As of today, this in fact the major source of KG alignment problems. Our proposed method for KG extension involves extending a reference KG through one or more candidate KGs. This implies that the reference KG typically possesses a more comprehensive and high-quality schema, serving as a foundation for KG extension. Our approach focuses on KGs that incorporate etypes associated with a substantial number of properties and complete schemas. As a result, we have selected the following cases: the bibliographic ontology dataset (BiblioTrack) \cite{euzenat2010results} and conference track (ConfTrack) \cite{zamazal2017ten} ($ra1$\footnote{https://owl.vse.cz/ontofarm/} version). From the bibliographic ontology dataset, we select \#101-103 and series \#301-304, which present real-life ontologies for bibliographic references from the web. We select the alignment between \#101 and \#304 as the training set for training our ML-based etype matcher, and the rest of the ontology alignments as the testing set. The conference track contains 16 ontologies, dealing with conference organizations, and 21 reference alignments. We set all 21 reference alignments from the conference track as the testing set to validate our etype matcher. Notice that we select the training and testing set from different datasets since we aim to prove the adaptation of our approach, which also prevents our approach from overfitting. 

For validating the algorithm of etype recognition, we build a dataset called EnType, since there is no publicly released dataset for such etype recognition task between two KGs. We exploit DBpedia infobox dataset\footnote{http://wikidata.dbpedia.org/services-resources/ontology} as the reference KG for providing reference etypes. Because DBpedia is a general-purpose KG that contains common etypes in the real world, where sufficient properties are applied for describing these etypes. Then we select candidate entities from DBpedia, SUMO
and several domain-specific datasets \cite{tang2008arnetminer}. 
The entities we selected mainly according to common etypes, more specifically, \textit{Person, Place, Event, Organization} and their sub-classes. Finally, we obtain 20,000 etype-entity candidate pairs, where 6,000 from DBpedia (EnType$_{Self}$) and 14,000 from the remaining resources (EnType$_{Gen}$). Need to notice, this dataset will be randomly separated into the training and testing set to implement the ML model.

\subsubsection{Feature Selection.}
Our approach applies a general binary classification strategy, which is independent of the specific ML model. Thus, the data label of positive and negative samples refers to if the pair etype-etype or etype-entity is aligned. Besides our property-based similarity metrics,  some of the most common string-based and language-based similarity metrics are selected as additional metrics for achieving better etype recognition performance. The data consists of three kinds of features, which are property-based similarity metrics ($Sim_H$, $Sim_V$ and $Sim_I$), string-based similarity metrics (N-gram \cite{euzenat2007ontology}, Longest common sub-sequence \cite{euzenat2007ontology}, Levenshtein distance \cite{yujian2007normalized}) and language-based similarity metrics (Wu and Palmer similarity \cite{palmer1994verb} and Word2Vec \cite{church2017word2vec}). These similarity metrics aim to measure different aspects of the relevance between the reference etype and candidates. Since all the above-mentioned similarity metrics are symmetric, the order of etype/entity in the candidate pair will not affect the final results. Moreover, we apply only property-based similarity metrics ($Sim_H$, $Sim_V$ and $Sim_I$) for etype recognition since the label of the candidate entity are commonly not relevant to its etype.

\subsubsection{Training Strategy.}
We have discussed that candidate pairs for applying binary classification in different etype recognition tasks. However, positive and negative samples in candidate pairs are usually not balanced in practice, more specifically, negative samples will be produced much more than positive samples. For instance, there will be one million candidate pairs generated when aligning two KGs that contain 1,000 etypes each, where only hundreds of them are positive samples. As a result, the ML model trained by such unbalanced datasets easily to be overfitted. To address this issue, we propose a model training strategy that increases the weight of positive samples to achieve a balanced training set and alleviate overfitting. By duplicating a part of positive samples, we achieve the best etype recognition performance by keeping the ratio of positive and negative samples as 1:10. Notice that such a data augmentation strategy is not applied to the testing set, the candidate pairs in the testing set are selected randomly to avoid interference.

\subsubsection{Dealing with Trivial Samples.}
In order to decrease the negative samples and avoid generating unnecessary candidate pairs, we prune the trivial samples which are obviously to be negative. Regarding different tasks, we have:
\begin{itemize}
    \item For etype alignment task, we apply label-based measurements to filter the obviously negative samples. Given two etypes $E_a$ and $E_b$ from a candidate pair, we define the pre-selection factor $PS_s$ as: 
\begin{equation}
PS_s = Ngram(E_a,E_b) + Word2Vec(E_a,E_b)
\label{equ:7}
\end{equation}
    where $Ngram(\cdot)$ and $Word2Vec(\cdot)$ are two similarity measurements; $Ngram(\cdot)$ assesses lexical similarity between strings by analyzing contiguous sequences of n items from a given string \cite{euzenat2007ontology}; $Word2Vec(\cdot)$ captures semantic similarity between strings through deep feature representations using the Word2Vec model \cite{church2017word2vec}. Thus, we consider $(E_a,E_b)$ to be an obviously negative candidate etype pair if $PS_s$ is greater than the threshold $th$. Experimentally, we find that $th = 0.3$ will lead to better results, where around 70\% of trivial sample pairs are reduced while maintaining minimal interference with the majority of positive samples.
    
    \item For the etype recognition task, the obviously negative candidate pair is identified when two entities $I_a$ and $I_b$ have no shared property. Thus, such candidate pairs will be pruned before inputting into the etype recognizer.
\end{itemize}

\noindent
The pre-selection of trivial samples will effectively reduce the run-time for model training. Moreover, such a training strategy will help to decrease the risk of overfitting by alleviating redundant samples and to improve the performance of KG alignment.

\subsubsection{Evaluation metrics.}
In our experiment, we exploit standard evaluation metrics including precision (\textbf{Prec.}), recall (\textbf{Rec.}) and $F_1$-measure (\textbf{$F_{1}$}\textbf{-m.}), and additional $F_{0.5}$-measure (\textbf{$F_{0.5}$}\textbf{-m.}) and $F_2$-measure (\textbf{$F_{2}$}\textbf{-m.}) \cite{pour2020results} to comprehensively validate our method and compare it with state-of-the-art methods. We form the etype recognition candidates as pairs, where each pair consists of a reference etype and a candidate etype/entity alignment. 
$F_{\beta}$-measures (\textbf{$F_{\beta}$}\textbf{-m.}) are defined as the harmonic mean of recall and precision:
\begin{equation}
F_{\beta}\textbf{-m.} = (1+\beta^2) * \frac{Prec.* Rec.}{\beta^2*Prec.+Rec.}
\label{equ:8}
\end{equation}
where $\beta$ is the configuration factor that allows for weighting precision or recall more highly if it is more important for the use case. And metrics \textbf{$F_{1}$}\textbf{-m.}, \textbf{$F_{2}$}\textbf{-m.} and \textbf{$F_{0.5}$}\textbf{-m.} are defined when $\beta = \{1, 2, 0.5\}$, respectively. We consider the $F_{\beta}$\textbf{-m.} to be the most relevant metrics for evaluation since it reflects both recall and precision.

\subsection{Etype Alignment}

\subsubsection{Qualitative analysis.}
Table \ref{tab2} provides representative examples to show the etype similarity metrics of candidate etype-etype pairs from \textit{cmt-confof} (former four rows) and \textit{cmt-conference} (latter four rows) in ConfTrack. Value box “Match" demonstrates if two etypes are referring to the same concept, where $\times$ refers to a positive answer. We find that the value of our property-based similarity metrics indeed capture the contextual similarity between relevant etypes, where aligned etypes output higher values (e.g., \textit{paper-contribution}), in turn, non-aligned etypes return lower values (e.g., \textit{person-document}). With a broad observation of the metric values, we consider the property-based similarity metrics $Sim_H$, $Sim_V$ and $Sim_I$ are valid for etype-etype pairs. 

\begin{table}[!t]
\centering
\caption{Representative samples of property-based similarity $Sim_V$, $Sim_H$ and $Sim_I$ on etype-etype pairs.}
  \label{tab2}

\begin{tabular}{|c|c|c|c|c|c|}
\hline

Etype$_{ref}$   & Etype$_{cand}$         & $Sim_V$ & $Sim_H$ & $Sim_I$ & Match\\ \hline
Paper       & Contribution         & 1      & 0.853  & 0.730 & $\times$ \\ \hline
SubjectArea & Topic               & 0.756  & 0.740  & 0.857 & $\times$ \\ \hline
Author & Topic                & 0.198  & 0.353  & 0.018 & \\ \hline
Meta-Review & Poster               & 0      & 0.312  & 0.262 & \\ \hline
Chairman    & Chair                & 1      & 0.559  & 0.554 & $\times$ \\ \hline
Person      & Person               & 1      & 0.970  & 0.678 & $\times$ \\ \hline
Person      & Document             & 0.02   & 0.06   & 0 & \\ \hline
Chairman    & Publisher            & 0      & 0.07   & 0.195 & \\ \hline
\end{tabular}

\end{table}

\subsubsection{Quantitative Evaluation.}

\begin{table*}[!t]
\centering
\caption{Quantitative comparisons on etype alignment.}
  \label{tab1}

\begin{tabular}{@{}lcccccccccc@{}}
\toprule
\multirow{2}{*}{\textbf{Methods}} & \multicolumn{5}{c}{\textbf{ConfTrack}} & \multicolumn{5}{c}{\textbf{BiblioTrack}}   \\ \cmidrule(l){2-6}  \cmidrule(l){7-11} & \multicolumn{1}{l}{\textbf{ Prec. }} & \multicolumn{1}{l}{\textbf{ Rec. }} & \multicolumn{1}{l}{ \textbf{$F_{0.5}$}\textbf{-m.} } & \multicolumn{1}{l}{ \textbf{$F_1$}\textbf{-m.} } & \multicolumn{1}{l}{ \textbf{$F_2$}\textbf{-m.} } & \multicolumn{1}{l}{\textbf{ Prec. }} & \multicolumn{1}{l}{\textbf{ Rec. }} & \multicolumn{1}{l}{ \textbf{$F_{0.5}$}\textbf{-m.} } & \multicolumn{1}{l}{ \textbf{$F_1$}\textbf{-m.} } & \multicolumn{1}{l}{ \textbf{$F_2$}\textbf{-m.} }\\ \midrule
FCAMap \cite{chen2019identifying}  & 0.680  & 0.625  & 0.668 & 0.651 & 0.635                                & 0.820  & 0.783  & 0.812 & 0.801 & 0.790                               \\
AML \cite{faria2013agreementmakerlight}    & 0.832  & 0.630 & 0.782 & 0.717 & 0.662      & 0.869  & 0.822 & 0.859 & 0.845 & 0.830                           \\
LogMap \cite{jimenez2011logmap}  & 0.798  & 0.592  & 0.746 & 0.680 & 0.624                                & 0.832  & 0.694  & 0.800 & 0.757 & 0.718  \\
Alexandre et. al. \cite{bento2020ontology} & 0.795 & 0.638  & 0.758 & 0.708 & 0.664                                & 0.827 & 0.786  & 0.818 & 0.806 & 0.794     \\ 
Nkisi-Orji et. al. \cite{nkisi2018ontology}  & \textbf{0.860}  & 0.514  & 0.758 & 0.643 & 0.559                                & 0.858  & 0.682  & 0.816 & 0.760 & 0.711  \\
LogMapLt \cite{jimenez2011logmap} & 0.716 & 0.554  & 0.676 & 0.625 & 0.580                                & 0.796 & 0.781  & 0.793 & 0.788 & 0.784     \\ \hline
ETA$_{XGBoost}$      & 0.827 & \textbf{0.676} & \textbf{0.792}  & \textbf{0.744}  & \textbf{0.702}    &  \textbf{0.870}   & \textbf{0.832} & \textbf{0.862}  & \textbf{0.851}  & \textbf{0.840}    \\
ETA$_{ANN}$    & 0.813   & 0.604 & 0.760 & 0.693 & 0.636                     & 0.797 & 0.811 & 0.799 & 0.803  & 0.808    \\

\bottomrule                    
\end{tabular}
\end{table*}

We apply two ML models to evaluate the validity of our proposed similar metrics on the etype alignment task, including XGBoost \cite{chen2016xgboost} and artificial neural network (ANN) classifier \cite{nath2021automated}, namely ETA$_{XGBoost}$ and ETA$_{ANN}$. We compared our work with state-of-the-art etype alignment methods\footnote{as most of them came out of previous OAEI evaluation campaigns}, including FCAMap \cite{chen2019identifying}, AML \cite{faria2013agreementmakerlight}, CNN-based ontology matching \cite{bento2020ontology}, word-embedding-based ontology alignment \cite{nkisi2018ontology}, LogMap and LogMapLt \cite{jimenez2011logmap}. We calculate mentioned evaluation metrics for a comprehensive comparison. Notice that we focus on the result of etype-etype alignment in this experiment.

Table \ref{tab1} shows the results of our approach with the different models mentioned above, compared with the results of state-of-the-art methods. Firstly, we can find our approach with different models produces different results. ETA$_{XGBoost}$ performs better than ETA$_{ANN}$ and also outperforms other methods on BiblioTrack and most of the cases in ConfTrack. The method by Nkisi-Orji et. al. \cite{nkisi2018ontology} leads the precision on ConfTrack, while it poorly performs on recall for both datasets. AML also shows competitive overall results compared to other state-of-the-art methods. Considering the average results of our approach with different models are performing better or close to the state-of-the-art, we can say that our approach surpasses the state-of-the-art competitors on the etype alignment task\footnote{All methods do not have significant differences in running times.}. Meanwhile, the promising overall performance produced by two different models indicates that our proposed similarity metrics and etype alignment approach are valid and adaptive.

\subsection{Etype Recognition of Entities}

\begin{table}[!t]
\centering
\caption{Representative samples of $Sim_V$, $Sim_H$ and $Sim_I$ on etype-entity pairs.}
  \label{tab:4}

\begin{tabular}{|c|c|c|c|c|c|}
\hline
$C_{ref}$   & $I_{cand}$        & $Sim_V$ & $Sim_H$ & $Sim_I$ & Match \\ \hline
Person       & MiltHinton       & 1      & 0.787  & 0.873 & $\times$ \\ \hline
Person      & Jadakiss          & 1      & 0.645  & 0.305 & $\times$ \\ \hline
Person      & Boston            & 0.264  & 0.173  & 0.041 & \\ \hline
Place       & Boston            & 0.720  & 1      & 0.433 & $\times$ \\ \hline
Place    & Jadakiss             & 0.128  & 0.093  & 0.148 & \\ \hline

Organization & MiltHinton       & 0.070  & 0.022  & 0.092 & \\ \hline

\end{tabular}
\end{table}

\begin{table*}[!t]
\centering
\setlength{\abovecaptionskip}{5pt}    
\setlength{\belowcaptionskip}{1pt}
\caption{Quantitative evaluation of etype recognition on dataset EnType.}
  \label{tab:3}

\begin{tabular}{@{}lcccccccc@{}}
\toprule
\multirow{2}{*}{\textbf{Methods}} & \multicolumn{4}{c}{\textbf{Self recognition}} & \multicolumn{4}{c}{\textbf{General recognition}}   \\ \cmidrule(l){2-5}  \cmidrule(l){6-9} & \multicolumn{1}{c}{Person} & \multicolumn{1}{c}{Org.} & \multicolumn{1}{c}{Place} & \multicolumn{1}{c}{Event} & \multicolumn{1}{c}{Person} & \multicolumn{1}{c}{Org.} & \multicolumn{1}{c}{Place} & \multicolumn{1}{c}{Event} \\ \midrule
Sleeman et. al. \cite{sleeman2015entity} & 0.633 & 0.509 & 0.618 & 0.712 & 0.365 & 0.274 & 0.420 & 0.323\\
Giunchiglia et. al. \cite{giunchiglia2020entity} & 0.594 & 0.582 & 0.672 & 0.651 & 0.480 & 0.494 & 0.563 & 0.501\\ 
ETR$_{XGBoost}$        & 0.850 & \textbf{0.795}  & 0.820 & \textbf{0.814} & \textbf{0.634} & \textbf{0.509} & \textbf{0.610}  & \textbf{0.514} \\
ETR$_{ANN}$        & \textbf{0.863} & 0.760 & \textbf{0.837}  & 0.808 & 0.589 & 0.435 & 0.497 & 0.486 \\

\bottomrule

\end{tabular}
\end{table*}

\begin{table*}[!t]
\centering
\setlength{\abovecaptionskip}{5pt}    
\setlength{\belowcaptionskip}{1pt}
\caption{Quantitative evaluation of etype recognition on different entity resolutions.}
\label{tab:5}
\resizebox{1\columnwidth}{!}{

\begin{tabular}{@{}lcccccccc@{}}
\toprule
\multirow{2}{*}{\textbf{Methods}} & \multicolumn{4}{c}{\textbf{Person}} & \multicolumn{4}{c}{\textbf{Organization}}   \\ \cmidrule(l){2-5}  \cmidrule(l){6-9} & \multicolumn{1}{l}{Athlete} & \multicolumn{1}{l}{MilitaryPerson} & \multicolumn{1}{l}{Artist} & \multicolumn{1}{l}{Comedian} & \multicolumn{1}{l}{MilitaryOrg.} & \multicolumn{1}{l}{Company} & \multicolumn{1}{l}{SportsClub} & \multicolumn{1}{l}{ReligiousOrg.} \\ \midrule
Sleeman et. al. \cite{sleeman2015entity} & 0.639 & 0.472 & 0.690 & 0.597 & 0.479 & 0.463 & 0.637 & 0.510 \\
Giunchiglia et. al. \cite{giunchiglia2020entity} & 0.658  & 0.429 & 0.736 & 0.522 & 0.512 & 0.487 & 0.624 & 0.482 \\  
ETR$_{XGBoost}$      & \textbf{0.756}   & \textbf{0.508}   & \textbf{0.760}  & 0.581   & \textbf{0.656}   & 0.487          & \textbf{0.719}  & \textbf{0.550}\\ 
ETR$_{ANN}$     & 0.712   & 0.507          & 0.732  & \textbf{0.659}  & 0.613   & \textbf{0.528}          & 0.702  & 0.491  \\
\bottomrule
\end{tabular}
}
\end{table*}

\subsubsection{Qualitative Analysis}

Table \ref{tab:4} presents some examples of similarity metrics between reference etypes and candidate entities from the dataset EnType. Similar to Table \ref{tab2}, the value box “Match" with $\times$ refers to that the candidate entity is classified as the reference etype. Similar to what happened in candidate etype pairs, we find the similarity metrics of aligned candidate pairs return much higher values than unaligned pairs. Thus, we conclude the proposed similarity metrics are also valid for measuring etype-entity pairs.

\subsubsection{Quantitative Evaluation}
As for the evaluation of the etype recognizer, we involve two subsets EnType$_{Self}$ and EnType$_{Gen}$ in this experiment. The subset EnType$_{Self}$ contains candidate entities and reference etypes from the same KG, i.e., \textit{self recognition}, which should be a relatively easier task. In turn, \textit{general recognition} denotes the recognition on subset EnType$_{Gen}$, where candidate entities are selected from other resources. Table \ref{tab:3} shows the results of etype recognition ($F_1$-measure), where we exploit the same models as in the etype alignment task, i.e., ETR$_{XGBoost}$ and ETR$_{ANN}$. Two state-of-the-art methods\cite{sleeman2015entity, giunchiglia2020entity} are also included in the experiment. Firstly, we can find that our methods ETR using the proposed property-based similarity metrics surpass two state-of-the-art methods significantly. For the self recognition group, ETR$_{XGBoost}$ performs better in two cases while ETR$_{ANN}$ also achieves competitive results. For general recognition cases, the XGBoost-based ETR method keeps its stable performance and surpasses all comparing methods. We achieve promising results in both cases, which prove the validity of our similarity metrics and the etype recognition approach. Meanwhile, we find that the precision of the general recognition case is lower than that of self recognition, which follows the difficulty of the two cases.

Considering that etype recognition performance is affected by entity resolutions, we apply an additional experiment for aligning entities with more specific etypes. We select four sub-classes of etype \textit{person} and \textit{organization} and their corresponding entities as candidate pairs, respectively. We keep comparing ETR$_{XGBoost}$, ETR$_{ANN}$ and the same state-of-the-art methods in this experiment. Table \ref{tab:5} presents the $F_1$-measure of the recognition results. We can find our methods still achieve better recognition performance than other methods in each case. Both two models obtain promising overall performance on specific etype recognition, where ETR$_{ANN}$ performs better on \textit{Comedian} and \textit{Company}, and ETR$_{XGBoost}$ leads the rest of the cases. The experimental results show that our metrics and approach can also be applied for specific instance-level etype recognition, which further supports the performance of KG extension algorithm.

\subsection{Ablation Study}
We demonstrate ablation studies in this section for validating the effectiveness of some specific components introduced in our KG extension framework.

\subsubsection{Effect of similarity metrics}
The first ablation study is to evaluate if each of the proposed property-based similarity metrics is effective. In this experiment, we test the backbones\footnote{trained by all three metrics ($Sim_V, Sim_H, Sim_I$)} (B) which were used in the etype alignment and recognition tasks, respectively. Based on the backbones, we also design a controlled group that includes models trained without one of the property-based similarity metrics (i.e. B-Sim$_V$, B-Sim$_H$ and B-Sim$_I$) and models trained without all metrics (i.e. B-L). If the backbones perform better than the corresponding models in the controlled group, we can quantitatively conclude that each of the property-based similarity metrics ($Sim_V, Sim_H, Sim_I$) contributes to the etype alignment and recognition tasks. Table \ref{tab:6} demonstrates the $F_1$-measure of each group. We apply ConfTrack for etype alignment and EnType$_{Gen}$ for etype recognition. Note that we select two models for both cases as Table \ref{tab:6} shows. We find that backbones perform better than models in the controlled group, especially for models trained without all metrics. Thus, we consider all property-based similarity metrics contribute to better recognition performance. Particularly, Sim$_V$ and Sim$_H$ significantly affect the performance of etype alignment cases, and Sim$_I$ affects etype recognition cases more.

\begin{table}[!t]
\centering
\setlength{\abovecaptionskip}{5pt}    
\setlength{\belowcaptionskip}{2pt}
\caption{Ablation study on property-based similarity metrics.}
\label{tab:6}
\begin{tabular}{@{}llccccc@{}}
\toprule
\multicolumn{1}{c}{Dataset} & \multicolumn{1}{c}{Model}  & Backbone & B-Sim$_V$ &B-Sim$_H$ & B-Sim$_I$   & B-L\\ \midrule
\multirow{2}{*}{ConfTrack} & ETA$_{ANN}$  & \textbf{0.713}    & 0.635     & 0.639    & 0.660  & 0.618     \\
\addlinespace & ETA$_{XGBoost}$                & \textbf{0.740}    & 0.648     & 0.655    & 0.694  & 0.632     \\ \hline
\multirow{2}{*}{EnType$_{Gen}$} & ETR$_{ANN}$ & \textbf{0.537}  & 0.327  & 0.391  & 0.309  & -    \\
\addlinespace & ETR$_{XGBoost}$                    & \textbf{0.559}  & 0.413  & 0.402  & 0.385  & -     \\\bottomrule
\end{tabular}
\end{table}

\subsubsection{Effect of constraint factor}
In section 4.1, we defined a constraint factor $\lambda$ for calculating the metric $Sim_H$. This study aims to statistically identify the value of $\lambda$. We apply the dataset ConfTrack and its two best-performed models. The value of $\lambda$ is set evenly from 0.1 to 1 by discrete points. We evaluate if this pre-set factor affects the final recognition performance and obtain the best value of $\lambda$ for generic etype recognition. Table \ref{tab:7} demonstrates the results, where we highlight both the best and second-best results. We can find that different values of $\lambda$ do affect the final etype recognition performance. And two models show a similar trend that the best value of $\lambda$ is close to 0.5. As a result, we assign $\lambda = 0.5$ to calculate metric $Sim_H$ in our experiments.

\begin{table}[!t]
\centering
\setlength{\abovecaptionskip}{5pt}    
\setlength{\belowcaptionskip}{2pt}
\caption{Ablation study on the constraint factor $\lambda$.The best and second-best results are highlighted in {\color[HTML]{FF0000} red} and {\color[HTML]{0070C0} blue}, respectively.}
\label{tab:7}
\begin{tabular}{@{}clccccccccc@{}}
\toprule
\multicolumn{1}{c}{Factor} & \multicolumn{1}{c}{Model} & 0.1 & 0.2 & 0.3 & 0.4 & 0.5 & 0.6 & 0.7 & 0.8 & 0.9 \\ \midrule
\multirow{2}{*}{$\lambda$} & ETA$_{ANN}$ 
& 0.613 & 0.638     & 0.670   & 0.678  & {\color[HTML]{FF0000} 0.712} & {\color[HTML]{0070C0} 0.685}  & 0.679 & 0.653  & 0.630 \\
\addlinespace & ETA$_{XGBoost}$   
& 0.662 & 0.677 & 0.714   & {\color[HTML]{0070C0} 0.727} &  {\color[HTML]{FF0000} 0.729} & 0.716 & 0.654 & 0.711 &  0.705  \\ \bottomrule
\end{tabular}
\end{table}

\section{Case Study}
\label{sec6.1}

This section aims to qualitatively analyze the KG extension performance of our proposed method by use cases. In the case of KG extension, we assume that there will be a reference KG extended based on one or more candidate KGs. We aim to simulate a real-world scenario, where people extend the general purpose KG by specific-domain KGs to enlarge its usability. We select the widely applied schema.org as the reference KG. For the candidate KGs, we introduce two specific-domain KGs Transportation\footnote{\url{https://carlocorradini.github.io/Trentino_Transportation}} and educationtrentino\footnote{\url{https://alihamzaunitn.github.io/kdi-educationtrentino}}. These two KGs are created for presenting local transportation and education, respectively. We selected these KGs because they provide very different examples in terms of the number of properties and etypes. Moreover, almost all their etypes labels are human understandable, which helps qualitative analysis. 

\begin{figure}[!t]
  \centering
  \includegraphics[width=0.7\linewidth]{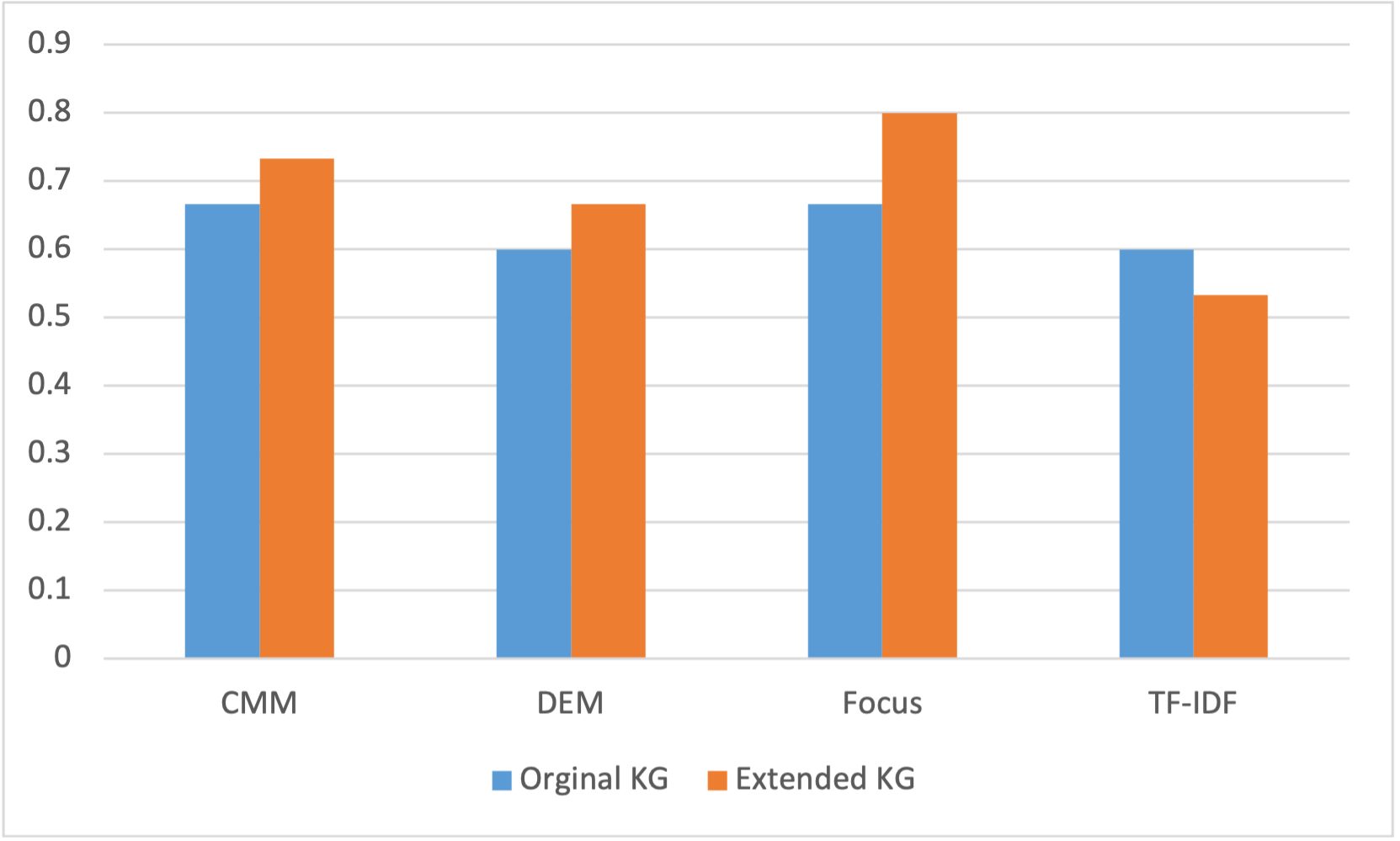}
  \caption{Experimental results of KG extension function.}
  \label{fig5}
\end{figure}

We introduce four ranking metrics in the case study, namely \textit{Class Match Measure (CMM)} \cite{alani2006metrics}, \textit{Density Measure (DEM)} \cite{alani2006metrics}, \textit{Focus} \cite{fumagalli2021ranking} and \textit{TF-IDF} \cite{salton1988term}. \textit{CMM} aims to evaluate the coverage of a KG for the given search etypes, by looking for etypes in each KG that have labels matching a search term either exactly or partially. \textit{DEM} is intended to approximate the representational density or information content of etypes and consequently the level of knowledge detail, considering etypes including the number of subclasses, the number of properties associated with that etype, the number of siblings, etc. \textit{Focus} aims to evaluate KG by identifying informative etypes with higher categorization relevance, by using the properties associated with the target etypes. \textit{TF-IDF} is also widely used for KG ranking by calculating relevance between potential KG with a specific term that describes the domain of interest. These metrics are used to evaluate the quality of the extended reference KG, where we compare the metrics on the original reference KG with the extended reference KG to see if the new-coming items from candidate KGs affect the quality of the reference KG. More specifically, we record the metrics for valid etypes in KG, and use the performance of these scores as a baseline, by selecting their scores for the top 15 etypes. The relevance of our approach is then measured in terms of accuracy (from 0 to 1) by checking how many etypes of ranking results are in the eypes ranking lists provided by the knowledge engineers. The output of this experiment is represented in Fig. \ref{fig5}.

The main observation is that our extended KG shows promising performance with all metrics. More specifically, with the given terms chosen by knowledge engineers, we can find the extended KG shows great categorization relevance by metric \textit{Focus}. The scores of \textit{CMM} and \textit{DEM} also show promising scores, respectively, which demonstrates the quality of the knowledge coverage and the representational density. Our KG gains a fair score on metric \textit{TF-IDF} since the reference schema.org is a general-purpose KG that contains pretty diverse concepts, which will affect the relevance with specific domain corpus to some extent. The second observation presents that the extended KG performs better than the original KG on most of the metrics. The increasing score of metric \textit{CMM} presents that the new-coming items enlarge the coverage of the etypes in the KG. The improving \textit{DEM} score shows the details of etypes from reference KG have been enhanced by the extended properties and subclasses. The increasing \textit{Focus} score also presents a trend where categories can be better identified from a taxonomy perspective. Notice that the extended schema.org has even more diverse concepts after KG extension, which becomes a possible reason for the slight decrease in \textit{TF-IDF} score. Overall, the experimental results in the case study demonstrate the ability of our method for KG alignment and extension.

\section{Related Work}
\label{sec7}
\subsection{Ontology Matching and Schema Alignment}
Ontology matching and schema alignment are attractive research topics in recent decades. In the early phases of the research, researchers mainly focused on string-based methods. String analysis techniques were defined including 1) string-based metrics (N-gram, Levenshtein, etc.), 2) syntactic operations (lemmatization, stop word removal, etc.) and 3) semantic analysis (synonyms, antonyms, etc.) \cite{cheatham2013string}. Sun et al, \cite{sun2015comparative} review a wide range of string similarity metrics and propose the ontology alignment method by selecting similarity metrics in different scales. Although string-based methods can lead to effective performance in many cases, selecting the right metric for matching specific datasets is the most challenging part. To solve this issue, an ensemble matching strategy is introduced in some studies \cite{bulygin2018combining,nezhadi2011ontology}, which apply multiple matchers based on different string-based metrics. The principle of these works is that the combined matchers are more powerful than individual ones. The structure of a KG has also been considered as important information for identifying etypes \cite{giunchiglia2012s,autayeu2010lightweight}. Such studies suppose that two etypes are more likely to be aligned if they have the same super-class or sub-class. The LogMap system \cite{jimenez2011logmap} uses a two-step matching strategy, that is, matches two etypes $E_a$ and $E_b$ by a lexical matcher, and then considers the etypes that are semantically close to $E_a$ are more likely to be semantically close to $E_b$. AML \cite{faria2013agreementmakerlight} introduces an ontology matching system that consists of a string-based matcher and a structure-based matcher, building internal correspondences by exploiting \textit{is-a} and \textit{part-of} relationships.

Machine learning techniques have been widely applied to this topic. Some studies model the etype matching task as a binary classification task, trying to encode the information like string and structure similarities as features for model training. Amrouch et al, \cite{amrouch2016decision} develop a decision tree model by exploiting lexical and semantic similarities of the etype labels to match schemas. By encoding the lexical similarity of the superclass and subclass as structure similarity, Bulygin and Stupnikov \cite{bulygin2019applying} improve the former method and achieve promising results. At the same time, formal concept analysis (FCA) lattices are applied in schema matching methods \cite{chen2019identifying,stumme2001fca}. To refine the health records searching outputs, Cure et al, \cite{cure2015formal} exploit FCA and Semantic Query Expansion to assist the end-user in defining their queries and in refining the expanded search space. Stumme et al, \cite{stumme2001fca} propose a bottom-up ontology merging approach by using FCA lattices to keep the ontology hierarchy. 

\subsection{Entity Type Recognition}
According to the different usage and motivation, studies on entity type recognition (also called entity typing) focus on three main directions: (1) recognizing the type of \textit{entity from text} \cite{xin2018improving,onoe2020fine}; (2) recognizing the type of entities from the single KG for \textit{KG completion} \cite{yaghoobzadeh2017multi,zhao2020connecting}; (3) recognizing the type of entities from different KGs for \textit{KG extension} \cite{sleeman2015entity}, which is the focus of this paper. Different from the former two tasks, we focus on recognizing the type of etypes/entities from other KGs for extending the reference KG automatically. In this field, some dedicated methods are proposed for specific datasets  \cite{dsouza2021towards, shalaby2016entity}. Rather than using label-based methods, some previous studies also consider properties as a possible solution,  \cite{sleeman2015entity,sleeman2013type} propose an etype recognition method by modeling etype recognition as a multi-class classification task. However, a pre-filtering step is needed since only properties shared across all candidates are counted for training and testing, which means there will be a few properties remaining after such filtering and a large amount of critical information will be discarded. Thus, the adaptation of such methods will be limited when applied in practice.  Giunchiglia and Fumagalli \cite{giunchiglia2020entity} propose a set of metrics for selecting the reference KG to improve the above method, which achieves improved performance with the support of a large amount of KGs. However, there are still limitations since these studies consider all properties with the same weight and neglect to distinguish properties that will contribute differently during etype recognition. 

\subsection{Knowledge Graph Extension}
KG extension aims to integrate additional knowledge from other KGs. It is different from KG self-completion which adds missing knowledge (concepts and properties) into the reference KG without exploiting other resources. In the context of the semantic web, most of the current cases locate the corresponding entity pairs in KGs and then directly integrate KGs by taking their union \cite{Hogan2020}. Several approaches for integrating schemas, given in terms of theories of classical first-order logic and rule bases, have been proposed. They either cast rules into classical logic or limit the interaction between rules and schemas \cite{Fahad2015, Steve1997}. Bruijn et al. \cite{Bruijn2011} presented three embeddings for ordinary and disjunctive nonground logic programs under the stable model semantics to select KGs for integration. Wiharja et al. \cite{Wiharja2018} improved the correctness of KG combinations based on Schema Aware Triple Classification (SATC), which enables sequential combinations of KG embedding approaches. However, these approaches are still limited when applied in practice. Novel methods for automatic schema integration are needed to extend general-domain KGs in a efficient and accurate way.

\section{Conclusions}
\nocite{1}
In this paper, we have proposed an ML-based framework for KG alignment and extension via a set of novel property-based similarity metrics. Firstly, we introduce a KG formalization method, which encodes etypes/entities and their corresponding properties into FCA contexts. We discuss that the corresponding properties are used to intentionally describe etypes, which provides us with a novel insight for identifying etypes. Then we propose three metrics for measuring the contextual similarity between reference etypes and candidate etypes/entities, namely the horizontal similarity $Sim_H$, the vertical similarity $Sim_V$, and the informational similarity $Sim_I$. Based on our proposed metrics, we introduce the framework with detailed algorithms and modules for KG alignment and extension. Thus, we validate our framework for the corresponding tasks. Compared with the state-of-the-art methods, the experimental results show the validity of the similarity metrics and the superiority of the proposed KG alignment methods, both quantitatively and qualitatively. Our future work will focus on exploring the further utilization of the proposed similarity metrics in tasks such as KG refinement and completion, including complex KG matching cases and resolving conflicts during the KG extension process. This exploration is expected to enhance the depth and accuracy of knowledge extraction and increase the practical applicability and adaptability of our proposed methods.

\section*{Acknowledgements}

The research conducted by Fausto Giunchiglia 
has received funding from the \emph{InteropEHRate} project, co-funded by the European Union (EU) Horizon 2020 program under grant number 826106, and the research conducted by Daqian Shi has received funding from the program of China Scholarships Council (No.  202007820024).



\bibliographystyle{splncs04}
\bibliography{refs}
\end{document}